%% file: main.tex
\documentclass[10pt,twocolumn,letterpaper]{article}

\usepackage{cvpr}              

\usepackage{soul}

\input{preamble}
\definecolor{cvprblue}{rgb}{0.21,0.49,0.74}
\usepackage[pagebackref,breaklinks,colorlinks,allcolors=cvprblue]{hyperref}

\def\paperID{} 
\def\confName{CVPR}
\def\confYear{2026}

\title{Metric-Guided Feature Fusion of Visual Foundation Models for\\Segmentation Tasks}

\author{
Yachan Guo\textsuperscript{1,2} \quad
Jose L. G\'{o}mez\textsuperscript{1,2} \quad
Danna Xue\textsuperscript{2} \quad
Yi Xiao\textsuperscript{3}\thanks{~Corresponding author.} \quad
Antonio M. L\'{o}pez\textsuperscript{1,2}\\[6pt]
\textsuperscript{1}Universitat Aut\`{o}noma de Barcelona \quad
\textsuperscript{2}Computer Vision Center \quad\\[6pt]
\textsuperscript{3}Harbin Institute of Technology, Shenzhen\\[4pt]
}

\begin{document}
\maketitle
\input{sec/0_abs}
\input{sec/1_introduction}
\input{sec/2_related_work}

\input{sec/3_method}
\input{sec/4_experiment}

\input{sec/5_conclusion}
\section*{Acknowledgements} 
This work was supported by the Spanish grant Ref. PID2024-157936NBI00 (HAMILTON project), funded by MICIU/AEI/10.13039/501100011033 and by FEDER, UE. A.M. López acknowledges financial support for his general research activities from ICREA under the ICREA Academia
Program. Y. Guo acknowledges support for her research from the China Scholarship Council through the Ph.D. Grant Ref 202208310071. All authors acknowledge the support of the Generalitat de Catalunya through the CERCA Program and its ACCIÓ Agency for CVC’s general activities.

{
    \small
    \bibliographystyle{ieeenat_fullname}
    \bibliography{main}
}

\input{sec/X_suppl}

\end{document}

%% file: sec/0_abs.tex
\begin{abstract}
Although large-scale visual foundation models (VFMs) achieve remarkable performance in semantic understanding, they still underperform in instance-aware dense prediction tasks. 
They exhibit different biases in representation: for instance, promptable segmentation models (\textit{e.g.}, SAM2) focus on fine-grained region boundaries, while self-supervised models (\textit{e.g.}, DINOv3) emphasize object-level structure. This observation highlights the potential of combining complementary features from different VFMs to enhance downstream dense prediction tasks. However, naive multi-VFM fusion seldom leads to reliable gains, and interpretable principles for leveraging their complementary features are still underexplored. 
In this work, we propose a metric-guided approach that effectively selects and aggregates complementary features from different VFMs based on explicit assessment scores. Specifically, we design a suite of label-free metrics in feature space across two aspects, \textbf{Structural Coherence} and \textbf{Edge Fidelity}, to assess features of VFM encoders. Guided by these scores, we identify complementary ``edge-strong'' and ``structure-strong'' encoder pairs, and integrate them via a master–auxiliary fusion scheme. 
This feature fusion requires no complex architectural changes and is trained only in a single stage. 
Our model shows \textbf{consistent performance gains} across multiple dense prediction tasks compared with the baselines, with better object-level semantics and more accurately localized boundaries.
The code is available at \url{https://github.com/gyc-code/metric-guided-fusion}. 

\end{abstract}

%% file: sec/1_introduction.tex
\section{Introduction}
\label{sec:intro}
Large-scale visual foundation models (VFMs) have become the \textit{de facto} starting point for a wide range of downstream vision tasks due to their remarkable scalability and transferability. Prominent examples include vision-language models such as CLIP~\citep{radford2021learning}, which 
enable open-vocabulary image classification and segmentation through text-image cross-modal alignment~\citep{wysoczanska2024clip,ilharco2021openclip,liang2023open}
; self-supervised Vision Transformers (ViTs) such as DINO and 
related variants
\citep{caron2021emerging,oquab2023dinov2,simeoni2025dinov3}, which achieve state-of-the-art semantic segmentation performance under challenging conditions like domain adaptation
\citep{wei2024stronger,guo2025uda4inst} 
; and promptable segmentation models such as SAM~\citep{kirillov2023segment} and SAM2~\citep{ravi2024sam}, whose encoder-decoder architecture produces real-time and high-quality region proposals that are well-suited for semantic segmentation and active learning~\citep{wei2024semantic, SAM4UDASS, Serrat2024}.


Given their strong performance in pixel-level semantic understanding, VFMs are naturally expected to excel across a broad range of dense prediction tasks. Instance segmentation is a representative example, where precise edge localization and instance-level disentanglement are critical.
Therefore, we aim to leverage 
VFMs with strong empirical validation and widespread adoption in recent literature~\citep{espinosa2024there,wang2024sam,li2024omg,ranzinger2024radio,heinrich2025radiov25,yu2024towards,wei2024stronger,wei2024semantic,yuan2024open,li2024segment}, such as DINO and SAM, to tackle these tasks effectively.

\begin{figure}
    \centering
    \includegraphics[width=\linewidth]{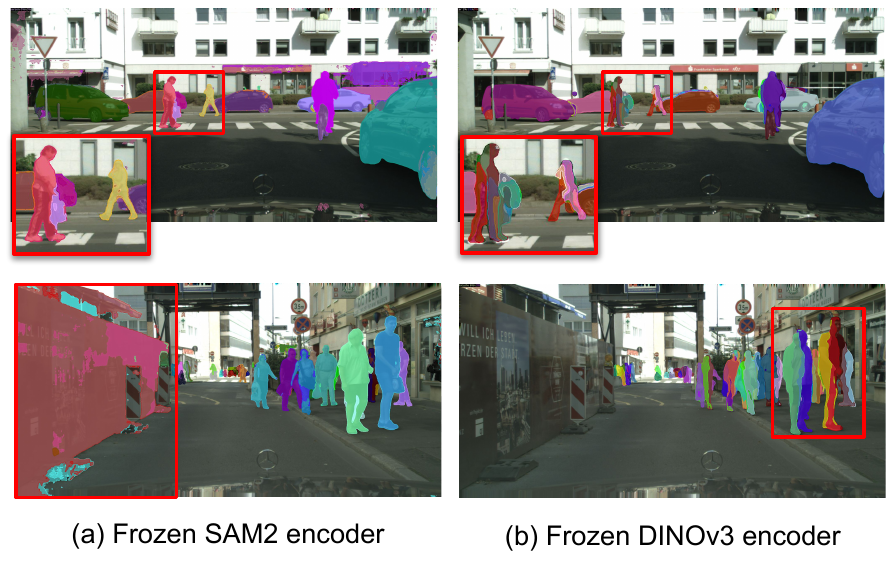}
    \caption{Failures of the frozen VFMs' encoders applied to instance segmentation: SAM-like models exhibit \emph{category confusions} in context-rich scenes; DINO tends to \emph{over-segment} a single object into multiple instances.}
    \label{fig:seg_problem}
    \vspace{-3mm}
\end{figure}

However, our preliminary study yields a counterintuitive observation. Despite being trained on vastly larger datasets, both DINO and SAM encoders significantly underperform the ImageNet-pretrained Swin Transformer when embedded in standard dense prediction pipelines under identical training protocols.
An example is illustrated in \cref{fig:seg_problem}: when VFMs' encoders are integrated with the decoder of Mask2Former~\citep{cheng2022masked} for instance segmentation, the predicted mask often suffers from either imprecise boundaries or over-segmentation, with DINO particularly prone to the latter. 
These findings underscore the need for further exploration to uncover VFMs' potential in dense prediction tasks. 

Recent studies indicate that pretrained ViT encoders exhibit systematic \emph{biases}~\citep{naseer2021intriguing, BaiECCV22HAT}: they tend to prioritize global shape representation, while underutilizing high-frequency details and edge cues unless explicitly supervised. 
For VFMs trained on larger and more diverse datasets, these biases are likely more complex and subtle than in ImageNet-pretrained models. Intriguingly, our preliminary qualitative analysis aligns with this hypothesis: when applied to an instance-aware prediction task, SAM features tend to emphasize fine object edges and boundaries, whereas DINO features favor smoother, structurally coherent regions 
. This observation suggests that different VFMs potentially capture complementary aspects of object representation, offering an opportunity for effective collaboration to improve dense prediction tasks. 

This raises two practical questions: 
(i) How can we \emph{assess}, \emph{without labels and at low cost}, the edge- or structure-wise feature bias of a given VFM encoder?
(ii) Given such an assessment, how can we \emph{leverage this insight} to improve 
performance across dense prediction tasks?
Motivated by these objectives, we propose a metric-guided fusion framework that leverages the complementary 
\textbf{\emph{edge–structural bias}} of VFMs.

Specifically, we design a suite of \emph{label-free, feature-level} metrics to assess VFM encoder features along the edge-structural spectrum, covering spatial coherence, boundary concentration, frequency characteristics, and edge sharpness.
Based on these metrics, we identify a complementary pair of VFM encoders: one biased toward structural coherence (\emph{``structure-strong"}), and the other toward edge fidelity (\emph{``edge-strong"}). These two encoders are integrated into a unified framework via a \emph{master-auxiliary} scheme, 
and collaborate using a feature fusion strategy during downstream task training. Based on the proposed metrics, our strategy is to select feature stages that achieve the highest scores in the edge-wise metric from the \emph{auxiliary} encoder to fuse into the \emph{master} encoder. This strategy effectively rebalances the edge–structural feature distribution of the \emph{master} encoder, yielding consistent performance gains across semantic and instance segmentation tasks. 

Our contributions are as follows:
\begin{itemize}
\item We design a suite of label-free, feature-level metrics that quantify the structure-edge representational bias of VFM encoders, enabling interpretable and task-agnostic encoder profiling without ground-truth labels.
\item We propose a metric-guided master--auxiliary fusion framework that selects complementary encoder features based on SC/EF scores, requiring only single-stage training and minimal architectural modification.
\item We demonstrate consistent improvements over single-encoder baselines across semantic, instance, and panoptic segmentation on Cityscapes, COCO, ADE20K, and KITTI-360, as well as under synthetic-to-real transfer with larger backbones.
\end{itemize}

%% file: sec/2_related_work.tex
\section{Related Work}
\label{sec:related}

\noindent\textbf{Representational Bias of VFMs.} Recent VFMs mainly adopt ViT backbones~\citep{kirillov2023segment,simeoni2025dinov3}, which can capture global semantic dependencies effectively when trained at scale~\citep{radford2021learning,ilharco2021openclip,wang2023image,simeoni2025dinov3,ravi2024sam,wang2024git}. By providing strong pretrained features, these models have become a strong foundation for downstream vision tasks.
However, their learned representations differ substantially depending on the pretraining objective, leading to inherent representational biases. 
For instance, DINO-family models~\citep{li2023mask,oquab2023dinov2,simeoni2025dinov3} rely on self-supervised learning and self-distillation, which encourage implicit clustering in feature space and thus favor more semantically coherent representations. In contrast, the SAM series~\citep{kirillov2023segment,ravi2024sam} is trained with mask-based objectives that explicitly emphasize object boundaries, leading to representations that are more boundary-sensitive.
In our work, we exploit these complementary structure- and edge-dominant biases to form a richer feature spectrum for downstream segmentation.

\noindent\textbf{VFMs for Dense Prediction Tasks.}
Dense prediction tasks like instance segmentation demand both precise boundary localization and robust semantic understanding.
Modern frameworks such as Mask2Former~\citep{cheng2022masked} pair transformer-based backbones~\citep{dosovitskiy2020image,Liu_2021_ICCV} with unified mask heads. Recent work adapts VFMs to these tasks via parameter-efficient tuning of a single encoder~\citep{chen2022vision,jia2022visual,gurav2025prompt,chen2023sam}.
As an alternative, some works focus on multi-encoder feature fusion that combines complementary towers via early/mid-level feature merging, late mask/score aggregation, or learned routing~\citep{zipyourclip-iclr2024, yuan2024open,shlapentokh2024region,wysoczanska2024clip}. Representative encoder pairs include \emph{CLIP\,+\,SAM}~\citep{zipyourclip-iclr2024}, \emph{SAM\,+\,DINO families}~\citep{shlapentokh2024region}, and \emph{CLIP\,+\,DINO}~\citep{wysoczanska2024clip}. However, these approaches typically require multi-stage training and complex integration modules, and offer limited insight into \emph{why} a given pair is effective. In contrast, we introduce label-free feature-level metrics that identify effective encoder interactions along the structure--edge spectrum, enabling lightweight single-stage fusion with interpretable guidance.

\noindent\textbf{Feature-level Assessment for VFMs.} Understanding the internal representational biases of VFMs is critical for their effective use, yet it remains underexplored. Prior assessments are often object-centric, attributing model failures to properties of specific objects and relying on ground-truth labels~\citep{zhang2024quantifying}. While such analyses can explain \emph{what} makes a task difficult, they reveal little about \emph{how} the model responds internally. Other works~\citep{cai2025computer, el2024probing, BaiECCV22HAT, naseer2021intriguing} probe frozen features without labels, but rarely provide a systematic quantitative framework for assessing the structure--edge trade-off in VFM encoders. Our work addresses this gap with a label-free, feature-level assessment toolkit grounded in classical vision principles such as clustering, gradients, and frequency cues.
We use feature-level metrics to reveal how a VFM is biased toward edge- or structure-dominant representations, and leverage them to build a lightweight, decoder-agnostic framework for multi-encoder feature fusion.

%% file: sec/3_method.tex
\begin{figure}
    \centering
    \includegraphics[width=1.0\linewidth]{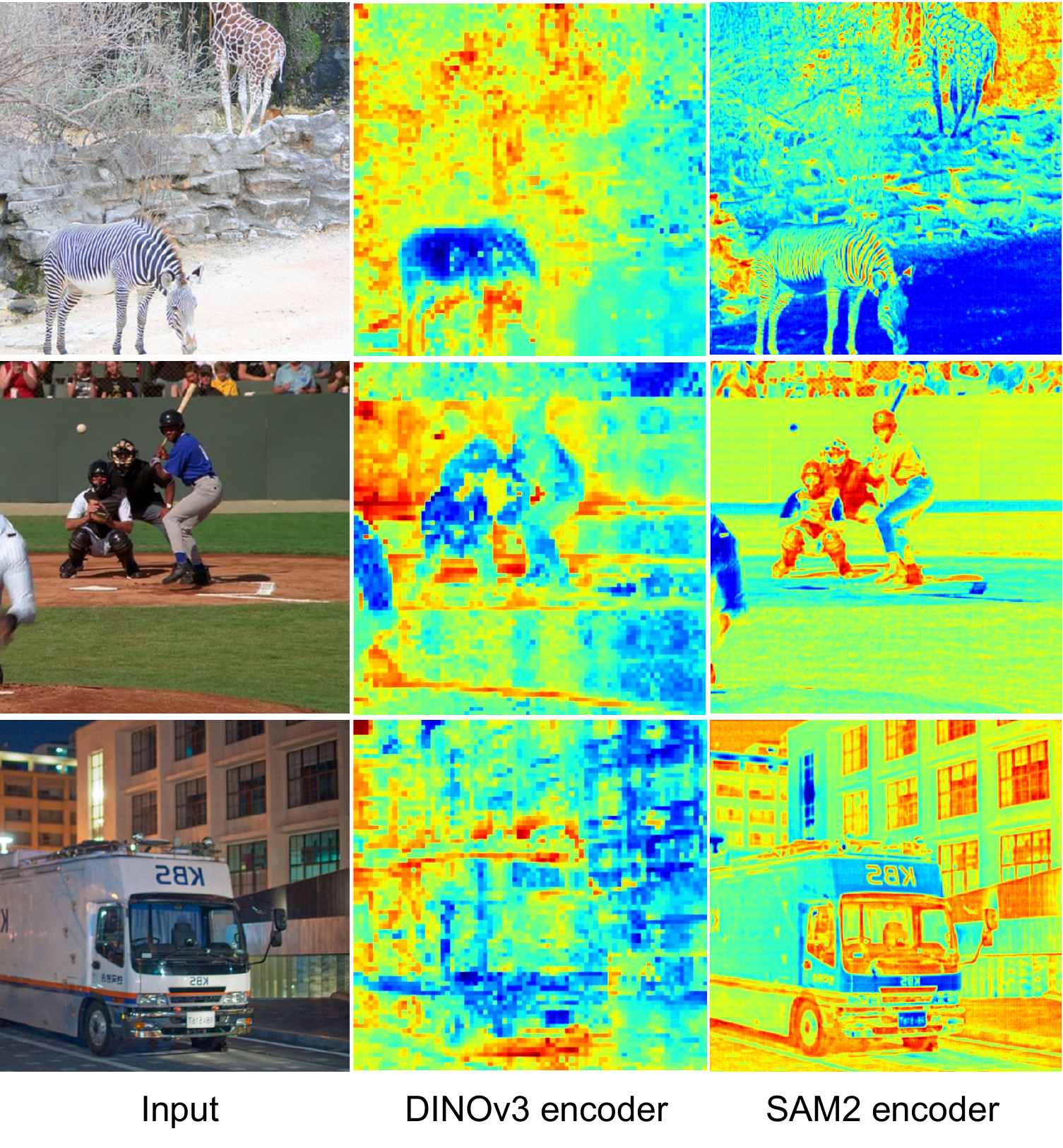}
   \caption{Channel-averaged feature activation maps (warm colors indicate high activation). DINOv3 activations spread across object bodies (structure-leaning), while SAM2 activations concentrate along boundaries (edge-leaning).}
    \label{fig:VFM_fusion_backbone_feature}
    \vspace{-3mm}
\end{figure}

\section{Method}
\subsection{Overview}
\begin{figure*}
    \centering
     \includegraphics[width=1\linewidth]{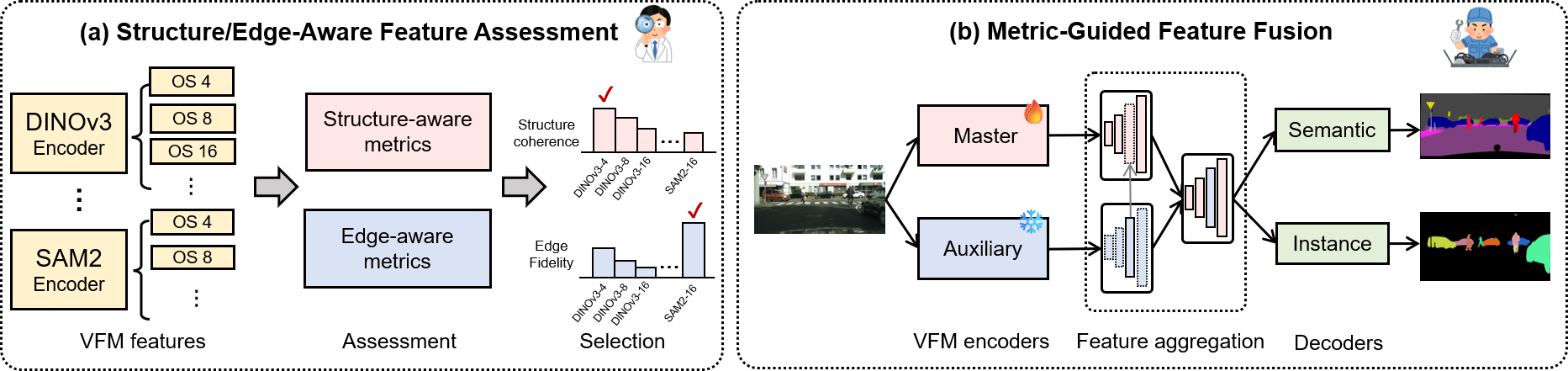}
     
    \caption{The overview of our framework. (a) Given features extracted by multiple VFM encoders, we design two categories of metrics, structure-aware and edge-aware, to evaluate features at different stages. Based on this assessment, we select the highest-scoring features and integrate their corresponding VFM encoders into our segmentation framework. (b) For the assessment-guided feature fusion, an input image is processed through a trainable master backbone (DINOv3-B) and a frozen auxiliary backbone (SAM2-B). Features are selected according to the assessment results, then aggregated and fed into decoder heads for different segmentation tasks.}
    \label{fig:VFM-framework}
\end{figure*}

Visual Foundation Models (VFMs) pretrained with different objectives exhibit distinct representation biases. As illustrated in Figure~\ref{fig:VFM_fusion_backbone_feature}, SAM2 concentrates activations along object boundaries, whereas DINOv3 distributes activations across object interiors. These biases align with two fundamental requirements of dense prediction: \textit{Structural Coherence} (SC), which ensures internally consistent features within objects and clear separation across them, and \textit{Edge Fidelity} (EF), which demands precise boundary delineation with minimal spatial leakage. To systematically characterize and leverage this complementarity, we propose a two-component framework, as illustrated in Figure~\ref{fig:VFM-framework}:
\textbf{(a) Structure/Edge-Aware Feature Assessment (Sec.~\ref{sec:assessments}):} We analyze multi-stage features from each VFM using label-free SC/EF diagnostics, producing per-stride assessments that identify complementary stages across encoders.
\textbf{(b) Metric-Guided Feature Fusion (Sec.~\ref{sec:strategy}):} Guided by these assessments, we construct a master-auxiliary fusion scheme that selectively injects complementary features at the metric-suggested stride, then feeds the fused multi-scale pyramid into task-specific decoder heads.

\subsection{Structure/Edge-Aware Feature Assessment}
\label{sec:assessments}

We propose two metrics considering the complementary aspects of VFM features: the SC metric quantifies how well structurally similar regions cluster in the feature space, while the EF metric measures the alignment of salient boundaries with respect to the input image. 
All metrics are computed without ground-truth labels, enabling their application to diverse VFM encoders trained across varying tasks. 

Given an input image $I$, a VFM encoder extracts the features $F^{(s)} \in \mathbb{R}^{C_s \times H_s \times W_s}$ at different encoder stages. Here, $s\in\mathcal{S}_{\mathrm{feat}}$ denotes the downsampling factor relative to the input image resolution (\textit{e.g.}, $\mathcal{S}_{\mathrm{feat}}=\{4,8,16,32\}$), and all metrics are computed using $F^{(s)}$.
More details are provided in the supplementary materials.

\subsubsection{Structural Coherence}
\label{sec:sac}
We assess SC from two complementary perspectives: whether features exhibit spatially coherent regions, and whether they form separable clusters in feature space.

\paragraph{Structured Feature Contrast (SFC).}
SFC measures the spatial consistency of features by comparing inter-patch contrast against intra-patch variability.
For each output stride $s$, we first compress the multi-channel feature map $F^{(s)}$ into a single-channel intensity map $M^{(s)}$, which is then divided into a $K{\times}K$ grid of equal-sized patches. For each patch $p$, we compute its mean $\mu_p$ and variance $\sigma_p^2$. The SFC is defined as:
\begin{equation}
\mathrm{SFC} = \frac{\mathrm{Var}(\{\mu_p\})}{\mathrm{Var}(\{\mu_p\}) + \mathrm{Mean}(\{\sigma_p^2\})}\in [0,1],
\end{equation}
where $\mathrm{Var}(\{\mu_p\})$ denotes the variance of patch means, capturing inter-patch structural contrast and $\mathrm{Mean}(\{\sigma_p^2\})$ denotes the average within-patch variance, reflecting local noise or texture complexity. 
High SFC indicates coherent, spatially distinct regions, whereas low SFC reflects dominant local fluctuations or near-uniform responses.

\noindent\textbf{Structural Clustering Score (SCS).}
SCS evaluates whether features form compact and well-separated groups in feature space. Given $F^{(s)}\in\mathbb{R}^{C_s\times H_s\times W_s}$, we flatten the spatial dimensions to obtain a pixel-wise feature matrix, then apply PCA for dimensionality reduction. Since the true number of semantic groups is unknown, we run k-means for a range of values $k\in K$ and compute the Silhouette coefficient~\citep{rousseeuw1987silhouettes} $\mathcal{S}_k$ for each. SCS is defined as:
\begin{equation}
\mathrm{SCS} = (\mathrm{median}(\{\mathcal{S}_k\}_{k \in K}) + 1)/2 \in [0,1].
\end{equation}
Taking the median across $K$ makes SCS robust to any single choice of $k$. Higher values indicate tighter, better-separated clustering.

\paragraph{Structural Coherence (SC).}
The final SC score is composed of the SFC and SCS scores to facilitate the feature selection in subsequent steps:
\begin{equation}
\mathrm{SC}=\sqrt{\mathrm{SFC}\cdot \mathrm{SCS}}.
\end{equation}


\subsubsection{Edge Fidelity (EF)}
\label{sec:ef}

Accurate segmentation requires precise boundary localization.
We introduce four metrics to assess whether VFM features faithfully capture image edges, organized into three aspects: \textit{spatial concentration} (Edge Concentration, Near-edge Concentration), which quantifies whether feature responses concentrate near image boundaries; \textit{frequency characteristics} (Frequency Content), which measures whether features preserve edge-typical frequency content; and \textit{spatial precision} (Spatial Precision), which evaluates whether edge responses are sharp rather than diffuse.

\paragraph{Preprocessing.}
We resize the input image to match the feature resolution $(H_s, W_s)$ and extract edge centerlines $E^{(s)} \in \{0,1\}^{H_s \times W_s}$ using the Sobel operator. Two scalar maps are derived from $F^{(s)}$: (i) a PCA projection $M_{\mathrm{PCA}}^{(s)}$ onto the first principal component, normalized to $[0,1]$, from which we compute a gradient magnitude map $G^{(s)} \in \mathbb{R}_{\geq 0}^{H_s \times W_s}$ via the Sobel operator; and (ii) a channel-wise L2 norm map $M_{\ell_2}^{(s)} = \|F^{(s)}\|_2$ along the channel dimension, followed by z-score normalization. Using morphological dilation with radii $r_{\text{in}}$ and $r_{\text{out}}$ ($r_{\text{in}} < r_{\text{out}}$), we define:
\begin{itemize}[leftmargin=*, itemsep=2pt]
    \item $A_{\text{in}} = \delta_{r_{\text{in}}}(E^{(s)})$: pixels within $r_{\text{in}}$ of edge centerlines.
    \item $A_{\text{near}} = \delta_{r_{\text{out}}}(E^{(s)}) \setminus \delta_{r_{\text{in}}}(E^{(s)})$: pixels in the narrow band between $r_{\text{in}}$ and $r_{\text{out}}$.
\end{itemize}
EC and NC operate on $G^{(s)}$ (derived from $M_{\mathrm{PCA}}^{(s)}$); FC and SP operate on $M_{\ell_2}^{(s)}$.

\paragraph{Edge Concentration (EC).}
EC measures how much of the total gradient energy falls within $A_{\text{in}}$:
\begin{equation}
\mathrm{EC} = \frac{\sum_{x \in A_{\text{in}}} G^{(s)}(x)}{\sum_{x} G^{(s)}(x)} \in [0,1].
\end{equation}
High EC indicates that feature responses are tightly localized around true boundaries.

\paragraph{Near-edge Concentration (NC).}
NC measures gradient energy in the band $A_{\text{near}}$ just outside the edge core. By excluding $A_{\text{in}}$ (already captured by EC), NC avoids redundantly rewarding edge-center responses and instead quantifies how well the near-edge region preserves boundary cues:
\begin{equation}
\mathrm{NC} = \frac{\sum_{x \in A_{\text{near}}} G^{(s)}(x)}{\sum_{x} G^{(s)}(x)} \in [0,1].
\end{equation}
High NC indicates clean edge responses without excessive spillover.

\paragraph{Frequency Content (FC).}
Edges predominantly occupy mid-to-high frequency bands, while smooth regions are dominated by low frequencies.
To quantify this, we apply a Hann window to $M_{\ell_2}^{(s)}$ to suppress spectral leakage, and compute the 2D power spectrum $P(u,v) = |\mathcal{F}(\cdot)(u,v)|^2$. Let $\rho$ denote the normalized radial frequency. FC measures the fraction of spectral energy above a low-frequency threshold $\rho_{\text{low}}$:
\begin{equation}
\mathrm{FC} = \frac{\sum_{\rho > \rho_{\text{low}}} P}{\sum_{\rho > 0} P} \in [0,1].
\end{equation}
High FC indicates that the features retain rich mid-to-high frequency content associated with edge details.

\paragraph{Spatial Precision (SP).}
SP assesses edge sharpness via a shift-sensitivity test. 
We translate $M_{\ell_2}^{(s)}$ by $r$ pixels along each of eight directions, computing the average normalized cross-correlation (NCC) with the original. The shift radii are sampled at log-uniform intervals up to a fraction of the spatial extent. Let $r_{\tau}$ be the minimum shift at which the average NCC drops below a threshold $\tau$. Sharp edges decorrelate quickly under small shifts, while blurry responses remain correlated over larger displacements:
\begin{equation}
\mathrm{SP} = 1 / (1 + \gamma \cdot r_{\tau}) \in [0,1],
\end{equation}
where $r_{\tau}$ is rescaled to image pixel units via the encoder stride, and $\gamma$ controls the decay rate. High SP indicates sharp, well-localized edge responses.


\paragraph{Composite EF Score.}
We combine the four metrics multiplicatively, so that weakness in any single aspect substantially reduces the overall score:
\begin{equation}
\mathrm{EF} = \alpha  \cdot \mathrm{EC}\cdot \mathrm{NC} \cdot \mathrm{FC}\cdot \mathrm{SP},
\end{equation}
where $\alpha $ is a scaling factor for numerical readability.

\subsection{Metric-Guided Feature Fusion}
\label{sec:strategy}

Based on the SC/EF metrics from Section~\ref{sec:assessments}, we design a metric-guided fusion architecture that integrates complementary VFM features.
As illustrated in Figure~\ref{fig:VFM-framework}(b), the framework consists of a \emph{master encoder} that provides the primary feature pyramid and an \emph{auxiliary encoder} that injects complementary features at a single stage.

\paragraph{Fusion Strategy.}
Given two encoders with complementary SC/EF properties, our fusion strategy is guided by the following principle: \emph{the master encoder is selected to match the primary requirement of the task, and the auxiliary encoder compensates for its relative weaknesses.}
Specifically, for segmentation tasks that demand strong semantic understanding, we select the encoder with higher SC scores as master and inject features from the auxiliary encoder at its highest EF stage, $s^* = \arg\max_s \mathrm{EF}_{\text{aux}}(s)$.
The fused feature pyramid is then constructed as:
\begin{equation}
F^{(s)} = \begin{cases}
F_{\text{aux}}^{(s)} & \text{if } s = s^* \\
F_{\text{master}}^{(s)} & \text{otherwise}
\end{cases}
\end{equation}

This strategy can generalize to other tasks. For instance, in edge-centric tasks, one may instead select the EF-dominant encoder as the master and inject SC-dominant features from the auxiliary. The overall framework builds upon Mask2Former~\citep{cheng2022masked}, with the sole modification being the encoder design. Accordingly, we use the SC-dominant DINOv3-B as the master encoder and the EF-dominant SAM2-B as the auxiliary encoder. The master encoder is set as trainable to adapt to downstream datasets, while the auxiliary encoder remains frozen to preserve its edge-aware representations.
The fused features are fed into a decoder with task-specific heads.

%% file: sec/4_experiment.tex
\section{Experiments}
\label{sec:experiments}

\subsection{Experimental Setup}

\noindent\textbf{Datasets.}
We train and evaluate on \textit{MS~COCO~2017}~\citep{lin2014microsoft} and \textit{Cityscapes}~\citep{cordts2016cityscapes} using official train/val splits.
We additionally validate on ADE20K (semantic), KITTI-360 (instance), and synthetic-to-real transfer (Urbansyn/Synscapes$\rightarrow$Cityscapes).

\noindent\textbf{VFM encoders.}
We consider four ViT-based VFM families:
SAM~\citep{kirillov2023segment} and SAM2~\citep{ravi2024sam} are pretrained with promptable mask supervision on SA-1B and extended video corpora;
DINOv2~\citep{oquab2023dinov2} and DINOv3~\citep{simeoni2025dinov3} are self-supervised on LVD-142M and LVD-1689M respectively.
To the best of our knowledge, none of these pretraining datasets explicitly include COCO, Cityscapes or other datasets we use. All encoders use Base-scale variants (ViT-B for DINO family, Hiera-B+ for SAM2) unless otherwise noted.
All encoders are paired with the same Mask2Former~\citep{cheng2022masked} decoder.

\noindent\textbf{Backbone features.}
Mask2Former requires a four-level feature pyramid at output strides OS~$\in\{4,8,16,32\}$. DINOv2/DINOv3 produce single-scale features, so we downsample them to form the required four-level pyramid. SAM2 (Hiera backbone) natively provides a four-scale hierarchy. 

\noindent\textbf{Training protocol.}
All experiments use a unified Mask2Former pipeline with AdamW optimizer (learning rate $10^{-4}$, weight decay $0.05$) and Warmup-PolyLR scheduler. 
We adopt Mask2Former purely as a training framework and initialize the segmentation head from scratch, so that all performance differences across encoders can be attributed to the backbone features rather than pretrained head weights.
We train for 368,750 iterations on COCO, 180,000 iterations on Cityscapes, Urbansyn, and Synscapes, and 160,000 iterations on ADE20K, all with batch size 3.
For input image resolution, we follow the default dataset-specific configurations provided by Mask2Former. Specifically, SAM and SAM2 use their official input resolution of $1024{\times}1024$.
We denote fully trainable encoders as ``FT'' and frozen encoders as ``FZ''.

\noindent\textbf{Metric hyperparameters.}
For SC metrics: we use a $K{=}16$ grid for SFC, and 32 PCA components with $k\!\in\!\{6,8,10\}$ for SCS.
For EF metrics: $r_{\text{in}}{=}3$, $r_{\text{out}}{=}7$ for EC/NC; $\rho_{\text{low}}{=}0.15$ for FC; $\tau{=}0.5$, $\gamma{=}\tfrac{1}{64}$ for SP; $\alpha{=}100$ for the EF composite score.
Sensitivity analysis in the supplementary confirms these choices are robust.

\noindent\textbf{Evaluation metrics.}
We report AP/AP$_{50}$/AP$_{75}$/AP$_s$ for instance segmentation, mIoU for semantic segmentation, and PQ for panoptic segmentation.

\begin{table}[t]
\centering
\setlength{\tabcolsep}{3pt}
\caption{COCO instance segmentation. `FT', `FZ' and `Hybrid' denote fine-tuned, frozen, and combining the fine-tuned master and the frozen auxiliary, respectively.}
\scalebox{0.85}[0.85]{
\begin{tabular}{l l c c c c c}
\toprule
Method & Backbone & Mode & AP & AP$_{50}$ & AP$_{75}$ & AP$_s$ \\
\midrule
Mask2Former~\citep{cheng2022masked}           & Swin-B   & FT     & 44.1 & 66.8 & 47.1 & 22.8 \\
Semantic-SAM~\citep{li2023semantic}         & Swin-T   & FT     & 46.1 &  --  &  --  & 27.1 \\
ViT-Adapter~\citep{chen2022vision}         & ViT-B    & FZ     & 41.8 & 65.1 & 44.9 &  --  \\
\midrule
SAM2                 & Hiera-B+ & FZ & 35.8 & 57.0 & 37.6 & 19.2 \\
DINOv3               & ViT-B    & FT     & 46.0 & 69.9 & 49.3 & 24.2 \\
Ours-D3S2                 & ViT-B    & Hybrid & \textbf{47.3} & \textbf{70.8} & \textbf{51.4} & \textbf{27.3} \\
\bottomrule
\end{tabular}
}
\label{tab:coco-apV2}
\end{table}

\begin{table}[t]
\centering
\caption{Cityscapes semantic and instance segmentation. Top: large-scale backbones under original protocols. Bottom: unified Mask2Former training with ViT-Base-scale encoders.}
\scalebox{0.84}[0.84]{
\begin{tabular}{l l c c c}
\toprule
Method & Backbone & Mode & AP & mIoU $\uparrow$ \\
\midrule
\multicolumn{5}{l}{\textit{Large-scale backbones}}\\
AM\mbox{-}RADIOv2.5~\citep{heinrich2025radiov25}    & ViT-g/14   & Hybrid & --   & 78.4 \\
SigLIP 2~\citep{tschannen2025siglip}                & ViT-g/16   & Hybrid & --   & 64.8 \\
OSM~\citep{yu2024towards}                          & ConvNeXt-L & FZ & --   & 80.2 \\
\midrule
\multicolumn{5}{l}{\textit{Unified Mask2Former protocol (ours)}}\\
Swin-B (IN21K)             & Swin-B     & FT     & 38.0 & 80.5   \\
DINOv3                        & ViT-B/16   & FT     & 35.6 & 81.2 \\
SAM2                          & Hiera-B+   & FZ & 35.8 & 79.7 \\
Ours-D3S2                          & ViT-B/16   & Hybrid & \textbf{39.5} & \textbf{82.8} \\
\bottomrule
\end{tabular}
}
\label{tab:cityscapes-comparisonV2}
\end{table}

\subsection{Method Performance and Comparison}
We evaluate the performance of our method on COCO and Cityscapes and compare it with SoTA segmentation methods \citep{cheng2022masked, li2023semantic, chen2022vision, heinrich2025radiov25, tschannen2025siglip, yu2024towards}
and the vanilla 
VFMs of DINOv3 and SAM2. 

\noindent\textbf{Instance Segmentation on COCO.}
Tab.~\ref{tab:coco-apV2} compares results on COCO validation set for instance segmentation.
Our model obtains a consistent improvement of AP$_{75}$ and AP$_s$ over the DINOv3 baseline, and achieves the higher AP, indicating superior localization accuracy and enhanced performance on small objects.
The gains in both AP and AP$_s$ are substantial, showing that the fusion strategy effectively incorporates the edge-strong features into the structure-strong master encoder.

\noindent\textbf{Semantic and Instance Segmentation on Cityscapes.}
\cref{tab:cityscapes-comparisonV2} shows that our model outperforms all compared methods in instance AP, consistent with the COCO results. For semantic segmentation, our method achieves competitive mIoU with ViT-Base-scale encoders while surpassing several much larger backbones (e.g., ViT-g) trained with specialized pipelines.

\noindent\textbf{Generalization across datasets and scales.}
Tab.~\ref{tab:extra_datasets_summary} demonstrates consistent improvements across ADE20K (semantic), KITTI-360 (instance), COCO (panoptic), and domain generalization with large-scale backbones. 

\begin{table}[t]
\centering
\caption{Generalization across datasets, tasks, and backbone scales. D3=DINOv3, S2=SAM2. Ours-D3S2: D3 (finetuned master) + S2 (frozen auxilury). Bottom rows use ViT-L-scale encoders with synthetic-to-real transfer.}
\label{tab:extra_datasets_summary}
\scalebox{0.82}{
\begin{tabular}{llccc}
\toprule
Dataset & Task (Metric) & D3 & S2 & Ours-D3S2 \\
\midrule
ADE20K & Semantic (mIoU) & 56.1 & 46.9 & \textbf{57.5} \\
KITTI-360 & Instance (AP) & 19.9 & 14.3 & \textbf{21.9} \\
COCO & Panoptic (PQ) & 55.6 & 43.7 & \textbf{56.9} \\
\midrule
\multicolumn{2}{l}{\textit{Large-scale (D3-L / S2-L)}} & & & \\
Urbansyn$\rightarrow$CS & Instance (AP) & 30.0 & 27.8 & \textbf{32.5} \\
Synscapes$\rightarrow$CS & Instance (AP) & 30.4 & 22.5& \textbf{33.4} \\
\bottomrule
\end{tabular}
}
\end{table}

\subsection{Characterizing and Exploiting Structure--Edge Bias with SC/EF Metrics}
\label{sec:analysis}

We investigate the structure--edge complementarity underlying our fusion strategy through five research questions, progressing from empirical characterization to label-free quantification, actionable design guidance, ground-truth validation, and training-time analysis.

\noindent \textbf{RQ1: Do VFMs exhibit distinct structure--edge biases?}
To ground the structure--edge hypothesis empirically, we compare four VFM families under a unified Mask2Former head on Cityscapes. Tab.~\ref{tab:per_class_ap} shows a clear category-level split: with frozen encoders, SAM2 performs best on boundary-sensitive categories, whereas DINO encoders lag; after fine-tuning, DINOv2 and DINOv3 improve markedly and dominate large rigid categories, while SAM2 remains relatively stronger on human and cyclist classes. SAM2 also benefits little from fine-tuning, suggesting a persistent edge-focused prior.
This trend is consistent across model generations, indicating that the observed bias reflects family-level pretraining tendencies rather than checkpoint-specific artifacts. Our fusion results further corroborate this finding: Ours-D3S2 (edge injection into structure-strong master encoder) yields the largest gains on boundary-sensitive categories, while Ours-S2D3 (structure injection into edge-strong master encoder) primarily improves large rigid objects. This directional pattern confirms that fusion selectively compensates for each encoder's weakness, consistent with the hypothesized structure--edge complementarity.

\begin{table}[t]
\centering
\setlength{\tabcolsep}{4pt}
\caption{Per-class instance AP on Cityscapes with Mask2Former and ViT-Base encoders.
S=SAM, S2=SAM2, D2=DINOv2, D3=DINOv3.
Ours-D3S2: DINOv3 (FT master) + SAM2 (frozen aux);
Ours-S2D3: SAM2 (frozen master) + DINOv3 (frozen aux).}

\label{tab:per_class_ap}
\scalebox{0.7}{
\begin{tabular}{l *{10}{c}}
\toprule
& \multicolumn{4}{c}{Frozen} & \multicolumn{4}{c}{Fine-tuned} & \multicolumn{2}{c}{Ours-D3S2} \\
\cmidrule(lr){2-5}\cmidrule(lr){6-9}\cmidrule(lr){10-11}
Class & S & S2 & D2 & D3 & S & S2 & D2 & D3 & Ours-D3S2 & Ours-S2D3 \\
\midrule
\multicolumn{11}{l}{\textit{Boundary-sensitive categories (high boundary-to-area ratio)}}\\
person & 28.9 & 35.1 & 11.9 & 12.4 & 34.0 & 34.7 & 26.1 & 30.4 & 34.4 & 33.6 \\
rider & 20.3 & 25.2 & 9.4 & 8.8 & 25.1 & 26.3 & 23.1 & 23.9 & 26.8 & 25.4 \\
mbike & 16.2 & 19.0 & 9.7 & 9.7 & 18.6 & 22.4 & 20.9 & 21.0 & 22.3 & 20.1 \\
bicycle & 16.6 & 19.9 & 9.3 & 7.6 & 19.9 & 20.9 & 17.2 & 18.5 & 20.1 & 18.6 \\
\midrule
\multicolumn{11}{l}{\textit{Large rigid categories}}\\
car & 52.5 & 56.6 & 38.1 & 38.7 & 56.7 & 48.3 & 48.8 & 52.7 & 55.3 & 55.6 \\
truck & 32.9 & 40.4 & 30.8 & 29.7 & 28.2 & 27.6 & 38.2 & 32.5 & 40.6 & 37.9 \\
bus & 51.6 & 59.3 & 49.6 & 49.3 & 49.7 & 46.3 & 60.8 & 60.5 & 64.6 & 61.0 \\
train & 21.9 & 31.2 & 33.0 & 38.7 & 24.8 & 42.4 & 48.7 & 45.1 & 48.3 & 45.0 \\
\midrule
Average & 30.1 & 35.8 & 24.0 & 24.3 & 32.1 & 33.6 & 35.5 & 35.6 & \textbf{39.1} & \textbf{37.2} \\
\bottomrule
\end{tabular}
}
\end{table}

\noindent \textbf{RQ2: Do SC/EF scores capture complementarity from pretrained features?}
RQ1 establishes the existence of structure--edge complementarity through downstream evaluation. However, this characterization requires task-specific training and labeled benchmarks. We now examine whether our label-free SC/EF metrics can reveal the same complementarity directly from pretrained encoder features, prior to any downstream adaptation.
Tab.~\ref{tab:sc_ef_profiles} presents label-free SC/EF scores computed on Cityscapes.
The SC/EF patterns clearly separate encoders along the structure--edge spectrum.
DINOv3 exhibits consistently high SC (0.53--0.73) and low EF (1.27--5.88), confirming its semantics-leaning behavior with limited boundary emphasis.
In contrast, SAM2 shows lower SC with a pronounced EF peak at OS=16 (17.13), revealing concentrated boundary sensitivity at specific stages.
Swin-B occupies an intermediate position with balanced SC and moderate EF across strides, as expected of a hierarchical Transformer backbone.
Crucially, these measurements are not merely descriptive---they identify actionable fusion targets.
The SC/EF contrast between DINOv3 and SAM2 confirms their complementarity, and SAM2's EF peak at OS=16 suggests that this stride offers the strongest boundary cues for injection, a prediction we validate next.

\begin{table}[t]
\centering

\caption{SC/EF metrics of frozen VFM encoders across output strides on Cityscapes. Higher SC indicates stronger semantic coherence; higher EF indicates stronger boundary sensitivity.}

\label{tab:sc_ef_profiles}
\resizebox{0.9\linewidth}{!}{%
\begin{tabular}{ccccccccc}
\toprule
Metric & \multicolumn{4}{c}{SC} & \multicolumn{4}{c}{EF} \\ 
\cmidrule(lr){2-5} \cmidrule(lr){6-9}
OS & 4 & 8 & 16 & 32 & 4 & 8 & 16 & 32 \\ \midrule
DINOv2 & 0.56 & 0.47 & 0.39 & 0.29 & 5.45 & 8.66 & 11.67 & 16.26 \\
DINOv3 & 0.73 & 0.71 & 0.65 & 0.53 & 1.27 & 1.64 & 2.33 & 5.88 \\
SAM & 0.60 & 0.51 & 0.43 & 0.29 & 6.30 & 7.10 & 9.91 & 13.96 \\
SAM2 & 0.49 & 0.44 & 0.11 & 0.41 & 6.60 & 8.59 & 17.13 & 12.47 \\
SWIN-B & 0.65 & 0.65 & 0.64 & 0.58 & 3.82 & 3.11 & 2.41 & 6.85 \\ \bottomrule
\end{tabular}%
}
\end{table}

\begin{figure}[t]
\centering

\begin{subfigure}{\linewidth}
  \centering
  \includegraphics[width=\linewidth]{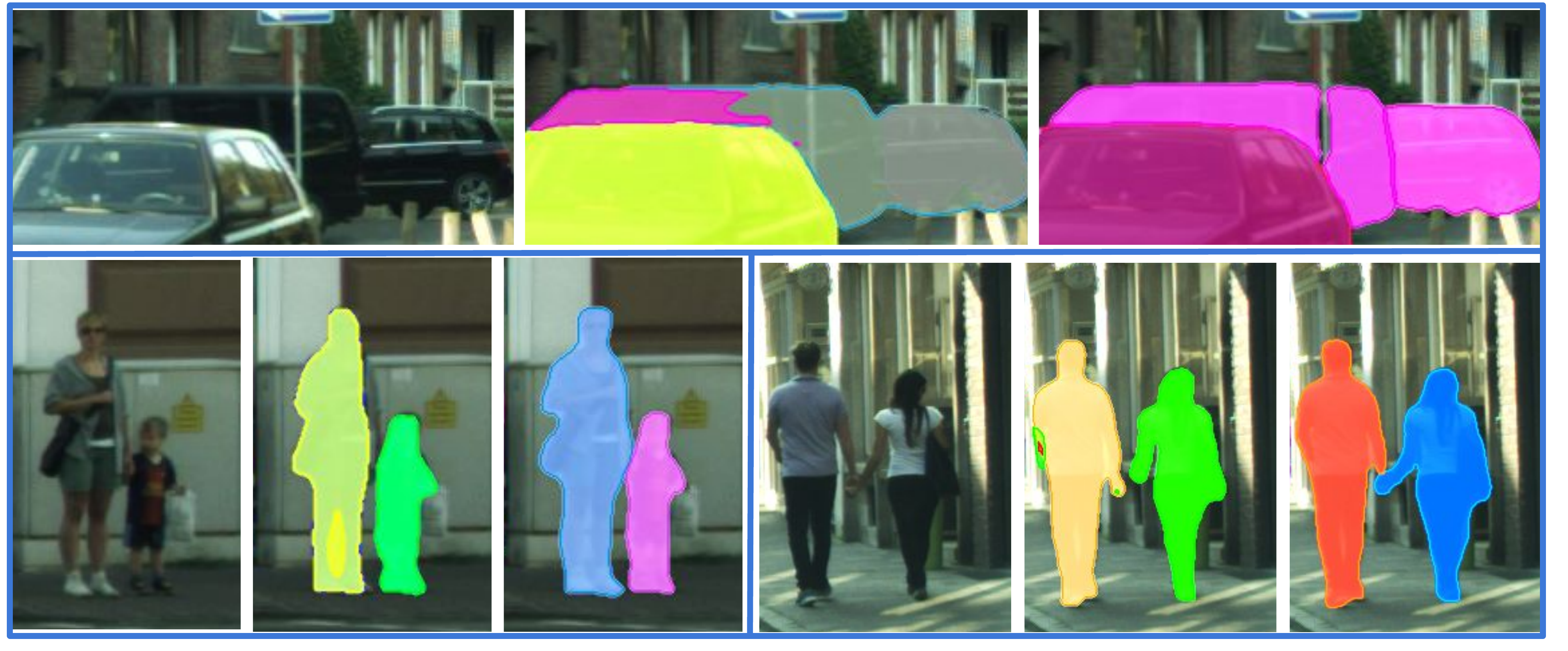}
  \caption{edge-enhanced fusion. Columns show \emph{RGB input}, \emph{finetuned DINOv3}, and \emph{Ours-D3S2} where SAM2 injects features into the DINOv3 master encoder.}
  \label{fig:qual_sam2_into_dino}
\end{subfigure}

\vspace{6pt}

\begin{subfigure}{\linewidth}
  \centering
  \includegraphics[width=\linewidth]{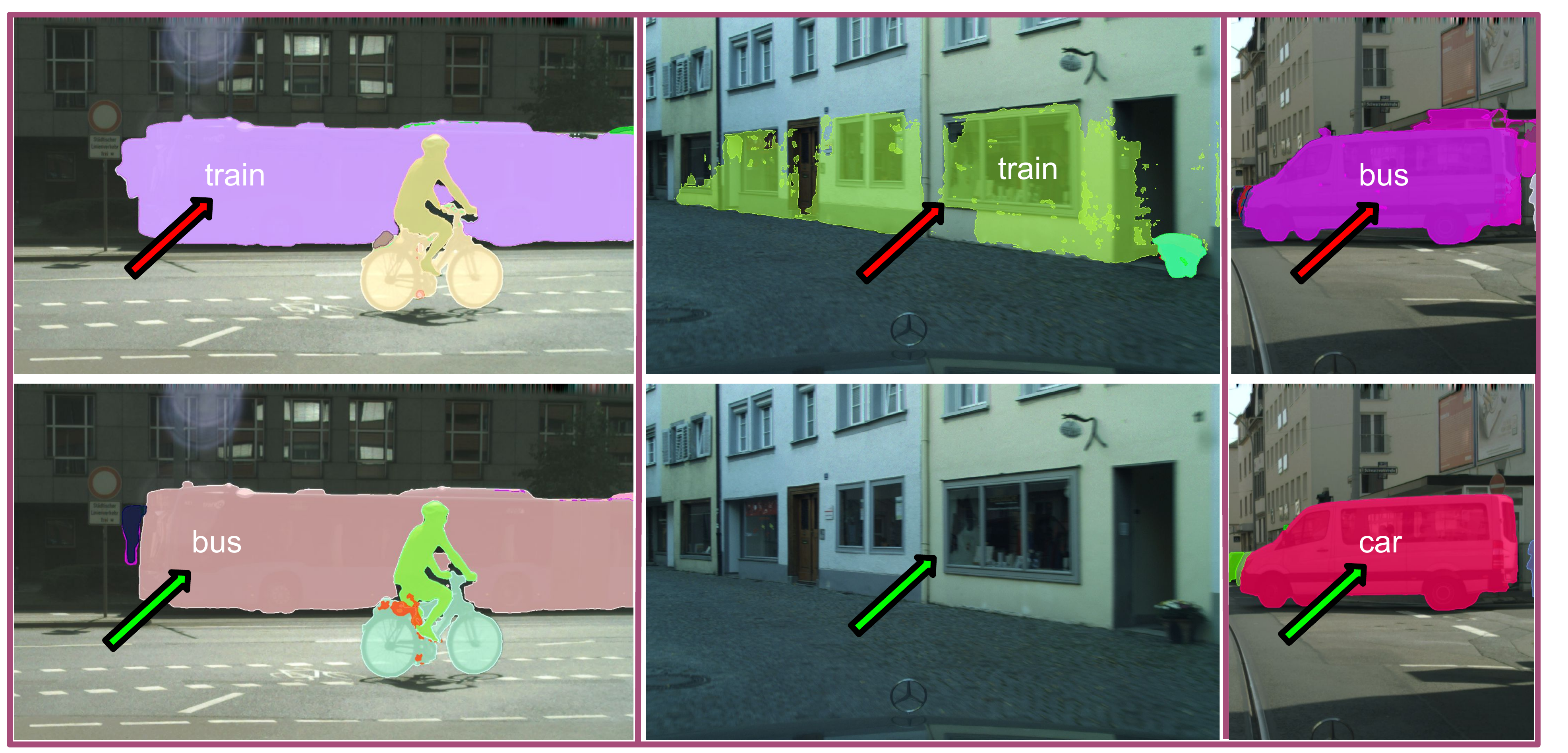}
  \caption{semantics-enhanced fusion. Rows show \emph{finetuned SAM2} and \emph{Ours-S2D3} where DINOv3 features are injected into the SAM2 master encoder.}
  \label{fig:qual_dinov3_into_sam2}
\end{subfigure}

\caption{Qualitative results on Cityscapes. 
(a) Injecting SAM2 features provides strong edge cues that sharpen DINOv3’s predictions. 
(b) Injecting DINOv3 features supplies richer semantic context and corrects SAM2’s predictions on large vehicles.}
\label{fig:qual_vfm_fusion}

\end{figure}

\noindent \textbf{RQ3: Can EF assessment predict the optimal injection stride?}
Since our fusion selects the SC-dominant encoder (DINOv3) as the main backbone, the auxiliary (SAM2) is responsible for injecting edge-aware features. The natural question is whether the auxiliary's EF score can identify the most effective injection stride. RQ2 shows that SAM2's EF peaks at OS=16; we now test whether this prediction holds in practice. In Tab.~\ref{tab:os_ablation}, we select $s^*=\arg\max_s \mathrm{EF}_{\text{aux}}(s)$ (underlined) for each main/auxiliary pair and compare against the best validation AP across all strides (bold). In three of four settings, the EF-suggested stride exactly matches the empirical optimum; in the remaining case, it is near-optimal and clearly outperforms early-stride injection. These results confirm that EF scores provide a practical alternative to exhaustive stride search. The same ablation also supports our training strategy: the combination of a fine-tuned main encoder with a frozen auxiliary (Setting~A) achieves the best overall AP, consistent with our design principle that the frozen auxiliary preserves its pretrained edge characteristics and keeps metric-guided selection stable. Per-class breakdowns for both DINOv2 and DINOv3 as master encoders are provided in the supplementary material.

\begin{table}[t]
\centering
\caption{Injection stride ablation on Cityscapes (instance AP). Underlined: EF-suggested stride ($s^*=\arg\max_s \mathrm{EF}_{\text{aux}}(s)$); bold: oracle best per row.}
\label{tab:os_ablation}
\scalebox{0.8}[0.8]{
\begin{tabular}{lcc|cccc}
\toprule
\multirow{2}{*}{Setting} & \multirow{2}{*}{master} & \multirow{2}{*}{Aux} &
\multicolumn{4}{c}{OS (AP)} \\
\cmidrule(lr){4-7}
 &  &  & 4 & 8 & 16 & 32 \\
\midrule
\multicolumn{7}{l}{\emph{DINOv3 as master; SAM2 as auxiliary}}\\
A & FT & FZ & 37.0 & 34.2 & \textbf{\underline{39.1}} & 35.3 \\
B & FZ & FZ & 27.1 & 32.0 & \textbf{\underline{35.4}} & 29.2 \\
\midrule
\multicolumn{7}{l}{\emph{SAM2 as master; DINOv3 as auxiliary}}\\
C & FT & FZ & 29.4 & 31.2 & \textbf{34.1} & \underline{32.2} \\
D & FZ & FZ & 33.9 & 34.7 & 35.8 & \textbf{\underline{37.2}} \\
\bottomrule
\end{tabular}
}
\end{table}


\noindent \textbf{RQ4: Do SC and EF capture meaningful feature properties?}
The preceding RQs provide indirect evidence that SC/EF scores are consistent with downstream performance. We further validate them against ground-truth measures on 1{,}000 Cityscapes images (Tab.~\ref{tab:metric_val}).
For SC, we compute a supervised counterpart $\text{SC}_{\text{GT}} = \sigma^2_{\text{inter}} / (\sigma^2_{\text{inter}} + \sigma^2_{\text{intra}})$ using semantic and instance labels. The unsupervised SC preserves the same encoder ranking as $\text{SC}_{\text{GT}}$ (Spearman $\rho=0.726$, $p<10^{-4}$).
For EF, we correlate per-stride EF scores with AP gains from injecting frozen SAM2 features into DINOv3. EF and $\Delta$AP follow the same trend (Spearman $\rho=0.80$), both peaking at OS=16. These results confirm that SC and EF can capture meaningful feature properties.



\begin{table}[t]
\centering
\caption{Validation of SC and EF on Cityscapes with ground-truth labels. (a) SC and $\text{SC}_{\text{GT}}$ per encoder, (b) EF and AP gains per stride.}
\label{tab:metric_val}
\subcaptionbox{SC validation, Spearman $\rho=0.726$, $p<10^{-4}$}{%
  \scalebox{0.8}{
    \begin{tabular}{lcc}
    \toprule
    Encoder & SC & $\text{SC}_{\text{GT}}$ \\
    \midrule
    DINOv3 & 0.75 & 0.46 \\
    Swin-B & 0.68 & 0.36 \\
    SAM2   & 0.56 & 0.09 \\ 
    \bottomrule
    \end{tabular}
  }}%
\hspace{30pt}
\subcaptionbox{EF validation, \textit{Spearman $\rho=0.80$}}{%
  \scalebox{0.8}{
    \begin{tabular}{lcc}
    \toprule
    OS & EF & $\Delta$AP \\
    \midrule
    4  & 6.21  & +2.8 \\
    8  & 8.82  & +7.7 \\
    16 & \textbf{17.89} & \textbf{+11.1} \\
    32 & 13.32 & +4.9 \\ 
    \bottomrule
    \end{tabular}
  }}
\end{table}

\noindent \textbf{RQ5: How does fine-tuning affect SC/EF?}
The preceding analyses rely on the features of pretrained VFMs. We now investigate whether downstream training reshapes these patterns, and what this implies for the frozen-auxiliary strategy.
Tab.~\ref{tab:profile_shift} tracks SC/EF evolution during fine-tuning on Cityscapes. 
For DINOv3, fine-tuning increases EF across strides while reducing late-stage SC, indicating a shift from structure-dominant to more balanced representations. For SAM2, the strong EF peak at OS=16 is greatly flattened while SC rises substantially at the same stage, showing that its edge-focused prior is weakened by task-specific training.
These shifts explain why freezing the auxiliary is effective: it preserves the pretrained pattern that metric-guided stride selection depends on. When the auxiliary is fine-tuned, its EF peak may shift or flatten, undermining the complementarity that motivated the fusion.

\begin{table}[t]
\centering

\caption{SC/EF values shifts after fine-tuning on Cityscapes. Values reported as ``before $\rightarrow$ after'' for each output stride.}

\label{tab:profile_shift}
\scalebox{0.8}[0.8]{
\begin{tabular}{llcc}
\toprule
Backbone & OS & SC & EF \\
\midrule
\multirow{4}{*}{DINOv3}
 & 4  & 0.73 $\rightarrow$ 0.73 & 1.27 $\rightarrow$ 1.65 \\
 & 8  & 0.71 $\rightarrow$ 0.67 & 1.64 $\rightarrow$ 3.51 \\
 & 16 & 0.65 $\rightarrow$ 0.55 & 2.33 $\rightarrow$ 6.24 \\
 & 32 & 0.53 $\rightarrow$ 0.44 & 5.88 $\rightarrow$ 8.79 \\
\midrule
\multirow{4}{*}{SAM2}
 & 4  & 0.49 $\rightarrow$ 0.43 & 6.60 $\rightarrow$ 9.67 \\
 & 8  & 0.44 $\rightarrow$ 0.39 & 8.59 $\rightarrow$ 11.35 \\
 & 16 & 0.11 $\rightarrow$ 0.46 & 17.13 $\rightarrow$ 9.30 \\
 & 32 & 0.41 $\rightarrow$ 0.46 & 12.47 $\rightarrow$ 9.30 \\
\bottomrule
\end{tabular}
}
\end{table}

%% file: sec/5_conclusion.tex
\section{Conclusion}
This work presents a metric-guided framework for leveraging complementary representational biases in visual foundation models for dense prediction. We proposed two label-free feature-space metrics, Structural Coherence and Edge Fidelity, to quantify the representational bias at every output stride across different encoders between semantic aggregation and boundary sensitivity. Guided by these metrics, we develop a straightforward master--auxiliary fusion strategy that injects complementary features at the stage guided by metrics, eliminating the need for complex architectural redesign or multi-stage training. Extensive experiments demonstrate consistent improvements across semantic, instance, and panoptic segmentation under unified evaluation protocols, with particularly notable improvements on boundary-sensitive categories where structure--edge complementarity is most critical. Overall, our findings establish feature-level profiling as a practical paradigm for interpreting pretrained encoders and informing their integration in downstream dense prediction tasks. We hope this work offers insights for understanding and leveraging representational bias in foundation models.

\section{Limitations and Future Work}
Despite these consistent gains, our framework has several limitations that warrant discussion. First, the proposed SC and EF metrics are manually designed, and their sensitivity can be influenced by implementation details such as patch granularity and frequency thresholds. Developing more adaptive mechanisms for quantifying representational bias remains a promising direction. Second, our current fusion strategy injects auxiliary features at a single stage. Since the optimal injection point likely depends on input solution, encoder scale, or downstream tasks, extending the framework to support adaptive, multi-stage fusion is a valuable avenue for future work. Third, our experiments focus on general-domain urban-scene benchmarks and a limited selection of VFM architectures. Broader validation across specialized domains, such as medical imaging and remote sensing, is necessary to confirm the applicability of our approach.

%% file: sec/X_suppl.tex
\clearpage
\setcounter{page}{1}
\maketitlesupplementary

\appendix
\section{Metric Details}

\subsection{Hyperparameter Sensitivity}
Our metrics involve several hyperparameters. To verify that the structure--edge characterization is robust to hyperparameters choices, we vary each within a range while keeping others fixed: SFC grid size $K\!\in\!\{8,12,16,20\}$; SCS PCA dimensions $\in\!\{16,32,64\}$ and cluster counts $k\!\in\!\{4,6,8,10,12\}$; EC/NC radii $r_{\text{in}}\!\in\!\{2,3,4\}$, $r_{\text{out}}\!\in\!\{6,7,8\}$; FC cutoff $\rho_{\text{low}}\!\in\!\{0.10,0.15,0.20\}$; SP threshold $\tau\!\in\!\{0.4,0.5,0.6\}$. In these cases, absolute metric values shift but the diagnosed structure--edge bias of each encoder remains consistent, confirming that the characterization is stable across reasonable hyperparameter choices.

\subsection{Cross-dataset consistency of SC/EF profiles.}
In the main paper,~\cref{tab:sc_ef_profiles} presents the SC and EF profiles of several VFMs on Cityscapes. 
The COCO results reported in~\cref{tab:coco_sc_ef_compact} exhibit a highly consistent pattern across datasets. 
All encoders preserve their bias on both the SC and EF axes, despite the substantial domain shift between COCO and Cityscapes. 
This indicates that the structure–edge characterization revealed by our assessment metrics reflects intrinsic properties of the pretrained VFM encoders rather than dataset-specific artifacts. 
Such cross-dataset stability further supports using these metrics to guide encoder pairing and OS selection in our fusion framework.







\begin{table}[h]
\centering
\caption{SC/EF profiles of frozen VFM encoders across output strides on COCO. Higher SC indicates stronger semantic coherence; higher EF indicates stronger boundary sensitivity.}

\label{tab:coco_sc_ef_compact}
\resizebox{0.9\linewidth}{!}{%
\begin{tabular}{ccccccccc}
\toprule
Metric & \multicolumn{4}{c}{SC} & \multicolumn{4}{c}{EF} \\ 
\cmidrule(lr){2-5} \cmidrule(lr){6-9}
OS & 4 & 8 & 16 & 32 & 4 & 8 & 16 & 32 \\ \midrule
DINOv2 & 0.66 & 0.62 & 0.55 & 0.47 & 1.95 & 3.37 & 5.30 & 8.09 \\
DINOv3 & 0.75 & 0.74 & 0.69 & 0.61 & 0.92 & 1.19 & 2.22 & 3.54 \\
SAM & 0.73 & 0.70 & 0.62 & 0.53 & 2.62 & 3.15 & 4.27 & 5.68 \\
SAM2 & 0.61 & 0.51 & 0.42 & 0.52 & 3.57 & 5.90 & 12.83 & 5.97 \\
Swin-B & 0.74 & 0.71 & 0.68 & 0.64 & 1.25 & 2.58 & 1.74 & 2.17 \\ \bottomrule
\end{tabular}%
}
\end{table}




\section{Additional Experimental Analysis}

\begin{table}[h]
\centering
\scriptsize
\setlength{\tabcolsep}{4pt}
\renewcommand{\arraystretch}{1.1}
\caption{Per-class AP on Cityscapes with frozen SAM2 as auxiliary. Each column injects SAM2 features at a single stride into the fine-tuned DINO main encoder. Bold: best stride per encoder.}
\label{tab:stage_swap_perclass}
\begin{tabular}{l c cccc c cccc}
\toprule
& \multicolumn{5}{c}{\textbf{DINOv2 (FT master)}} & \multicolumn{5}{c}{\textbf{DINOv3 (FT master)}} \\
\cmidrule(lr){2-6}\cmidrule(lr){7-11}
Class & base & OS4 & OS8 & OS16 & OS32 & base & OS4 & OS8 & OS16 & OS32 \\
\midrule
person  & 26.1 & 31.1 & 32.4 & 35.8 & 27.9 & 30.4 & 30.3 & 26.6 & 34.4 & 29.2 \\
rider   & 23.1 & 25.7 & 27.7 & 26.0 & 22.5 & 23.9 & 24.7 & 22.5 & 26.8 & 22.9 \\
car     & 48.8 & 55.0 & 54.6 & 56.2 & 50.4 & 52.7 & 53.6 & 49.4 & 55.3 & 50.7 \\
truck   & 38.2 & 38.3 & 37.9 & 36.8 & 35.8 & 32.5 & 38.6 & 36.9 & 40.6 & 36.0 \\
bus     & 60.8 & 63.4 & 62.5 & 63.0 & 60.4 & 60.5 & 59.2 & 56.4 & 64.6 & 59.7 \\
train   & 48.7 & 36.7 & 42.6 & 47.4 & 44.2 & 45.1 & 50.2 & 46.8 & 48.3 & 48.2 \\
mbike   & 20.9 & 23.2 & 22.0 & 23.1 & 21.9 & 21.0 & 19.9 & 19.7 & 22.3 & 19.3 \\
bicycle & 17.2 & 20.2 & 20.0 & 21.6 & 17.5 & 18.5 & 19.3 & 15.6 & 20.1 & 16.7 \\
\midrule
mAP     & 35.5 & 36.7 & 37.5 & \textbf{38.7} & 35.1 & 35.6 & 37.0 & 34.2 & \textbf{39.1} & 35.3 \\
\bottomrule
\end{tabular}
\end{table}

\begin{figure*}[t]
    \centering
    \includegraphics[width=0.82\linewidth]{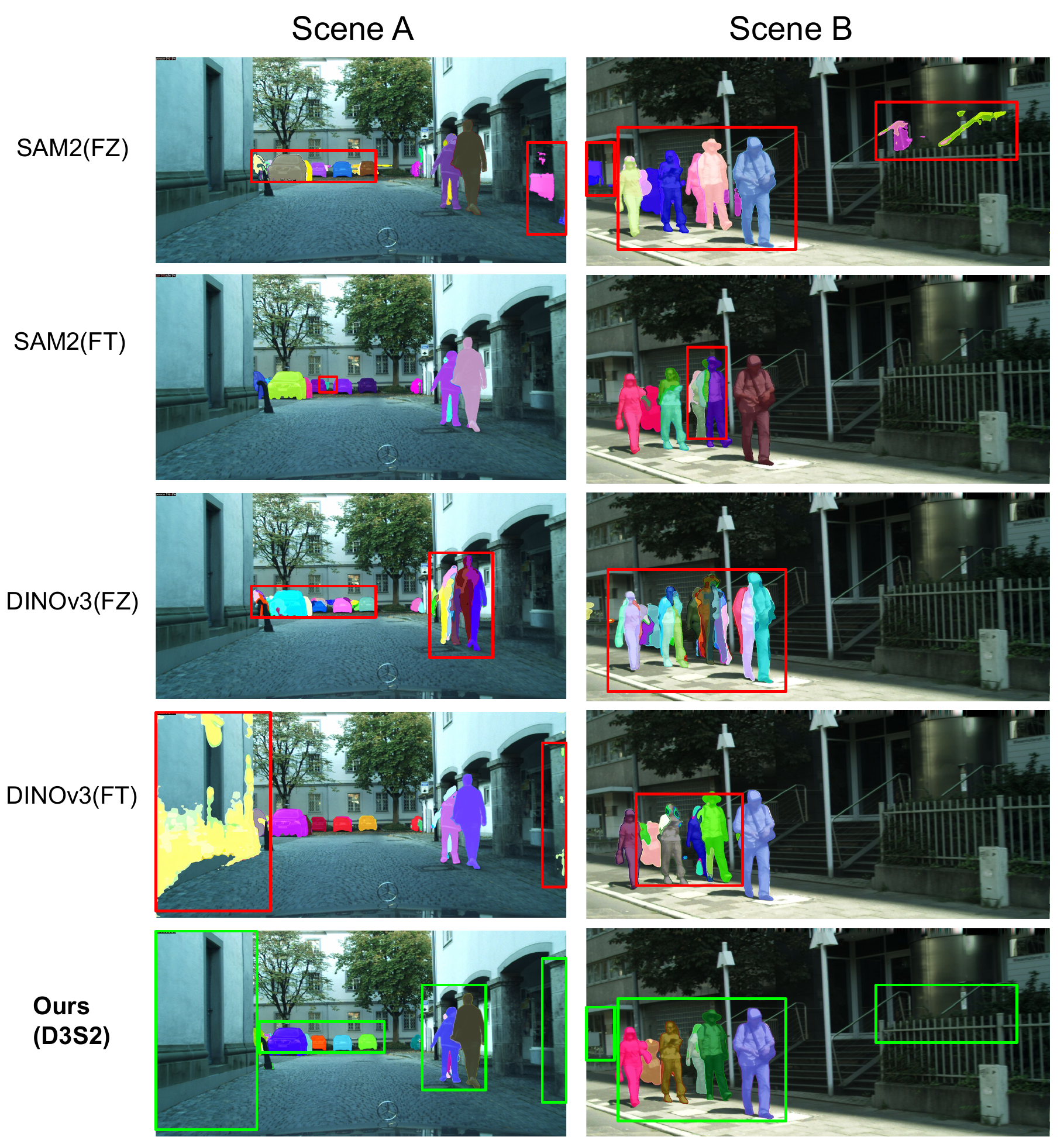}

    \caption{Qualitative comparison on Cityscapes. All outputs are unfiltered (no confidence thresholding). Red boxes: failure cases of individual encoders; green boxes: corresponding regions in ours-D3S2). Each column shows a different scene; rows show different encoder configurations.}
    
    \label{fig:qual_vfm_fusion}
\end{figure*}

\subsection{Per-class Analysis of Injection Stride}

\cref{tab:stage_swap_perclass} provides per-class AP on Cityscapes when injecting a single frozen SAM2 feature stage into a fine-tuned DINOv2 or DINOv3 master backbone, complementing the summary in the main paper (\cref{tab:os_ablation}). For both master encoders, injection at OS=16 yields the best average AP (38.7 for DINOv2, 39.1 for DINOv3), with gains most pronounced on boundary-sensitive classes such as person (+9.7/+4.0), bicycle (+4.4/+1.6), and motorcycle (+2.2/+1.3). Large rigid categories also benefit (e.g., car +7.4/+2.6), while injection at other strides produces smaller or inconsistent improvements. This confirms that the benefit is stage-specific and aligns with SAM2's EF score (\cref{tab:sc_ef_profiles}), which peaks at OS=16.

\subsection{Efficiency and Computational Cost}
\cref{tab:main_performance_cost} reports the computational cost of each backbone. Ours-D3S2 adds 774 GFLOPs over DINOv3-B (1857 vs.\ 1083) because we attach SAM2's backbone as an auxiliary edge provider. Despite the additional encoder, throughput remains comparable to single-encoder baselines such as DINOv2-B (1.4 vs.\ 1.6 img/s).


\begin{table}[t]
\centering
\small
\setlength{\tabcolsep}{2.8pt}
\renewcommand{\arraystretch}{1.0}
\caption{Computational cost at Cityscapes resolution ($1024\times2048$) with a Mask2Former head, measured on a single A40 GPU.}
\label{tab:main_performance_cost}
\begin{tabular}{l  c c c r}
\toprule
\textbf{Backbone} & \textbf{Params} & \textbf{GFLOPs} & \textbf{Throughput}\\
\midrule
DINOv2  & 108.1  & 1294 & 1.6 \\
DINOv3  & 107.1  & 1083 & 2.2 \\
SAM    & 111.1  & 2302 & 1.4 \\
SAM2    & 89.1  & 1132 & 2.3 \\
Ours-D3S2     & 176.2  & 1857 & 1.4 \\ 
\bottomrule
\end{tabular}
\end{table}

\section{Additional Qualitative Results}
\cref{fig:qual_vfm_fusion} extends the main-paper comparison (Fig.~1) by including fine-tuned variants and our fused model. The two VFM families exhibit distinct failure patterns under both frozen and fine-tuned settings. Frozen SAM2 preserves tight boundaries but suffers from \emph{category confusions}, missed small instances, and spurious fragments (e.g., distant regions in Scene~A, staircase artifacts in Scene~B). Frozen DINOv3 produces more coherent regions but tends to \emph{over-segment} single objects into multiple instances (e.g., fragmented person in Scene~A, split pedestrians in Scene~B). Fine-tuning SAM2 slightly improves recall but introduces noisier masks; fine-tuning DINOv3 alleviates some fragmentation but causes new artifacts such as mask spillover into background regions (Scene~A). Our fused model (D3S2) alleviates both failure modes by injecting edge-aware features from frozen SAM2 into the DINOv3 main encoder, yielding cleaner, better-separated, and more accurately localized predictions across both scenes.